\documentclass{article}

\usepackage[preprint,nonatbib]{neurips_2022}

\usepackage[T1]{fontenc}    
\usepackage{hyperref}       
\usepackage{url}            
\usepackage{booktabs}       
\usepackage{amsfonts}       
\usepackage{nicefrac}       
\usepackage{microtype}      
\usepackage{xcolor}         
\usepackage{graphicx}
\usepackage{subfig}
    \makeatletter

\newcommand*\samethanks[1][\value{footnote}]{\footnotemark[#1]}

\title{Docent: A content-based recommendation system to discover contemporary art}

\author{%
Antoine Fosset$^{1}$\thanks{Equal contribution},
Mohamed El-Mennaoui$^{1}$\samethanks,
Amine Rebei$^{1}$\samethanks$^{ }$$^{ }$$^{ }$
\thanks{Corresponding author \texttt{amine.rebei@docent-art.com}},\\
\textbf{Paul Calligaro$^{1}$,
Elise Farge Di Maria$^{1}$,
Hélène Nguyen-Ban$^{1}$},\\
\textbf{Francesca Rea$^{1}$,
Marie-Charlotte Vallade$^{1}$},\\
\textbf{Elisabetta Vitullo$^{1}$,
Christophe Zhang$^{1}$},\\
\textbf{Guillaume Charpiat$^{23}$,
Mathieu Rosenbaum$^{14}$}\\
$^{1}$Docent Art, $^{2}$INRIA Saclay, $^{3}$LISN, Université Paris-Saclay, $^{4}$CMAP, Ecole Polytechnique
}

\begin{document}

\maketitle

\begin{abstract}
Recommendation systems have been widely used in various domains such as music, films, e-shopping, etc. After mostly avoiding digitization, the art world has recently reached a technological turning point due to the pandemic, making online sales grow significantly as well as providing quantitative online data about artists and artworks. In this work, we present a content-based recommendation system on contemporary art relying on images of artworks and contextual metadata of artists. We gathered and annotated artworks with advanced and art-specific information to create a completely unique database that was used to train our models. With this information, we built a proximity graph between artworks. Similarly, we used NLP techniques to characterize the practices of the artists and we extracted information from exhibitions and other event history to create a proximity graph between artists. The power of graph analysis enables us to provide an artwork recommendation system based on a combination of visual and contextual information from artworks and artists. After an assessment by a team of art specialists, we get an average final rating of $75\%$ of meaningful artworks when compared to their professional evaluations.

\end{abstract}

\section{Introduction}

Human taste is inherently subjective, questioning the very possibility for it to be rationalized into a system, a challenge many have accepted in the content industry. Active research fields include music, films, images, videos and content at large. News outlets started including recommendation systems in their digital platforms to offer tailor-made news to their users and thus increase the user engagement\cite{Raza2022}. This evolution has enabled the emergence of major tech companies. Spotify, Netflix, Instagram, TikTok have developed powerful recommendation tools and used them to become leaders in their respective sectors.

Contrary to other creative industries, up until the restrictions brought on by the Covid-19 pandemic, most art world operators heavily relied on the offline. The art world was then pushed to an accelerated digitalization\cite{Noehrer2021}: to survive economically, galleries, fairs, and institutions have been led to invest in digital tools — websites, Online Viewing Rooms (OVRs), social media presence, etc. However, these options set aside any taste prediction. Art collectors who go on OVRs may feel overwhelmed by the quantity of artworks to browse before finding the right one.

Collectors rely on cultural prescription to identify and choose artists they wish to acquire. Social interactions, professional advice from an art advisor, traditional and social media all play a role in guiding a collector. However, such prescription is no longer efficient in the digital world, and remains exclusive, biased, and unpersonalized.
Reproducing the human eye of an art expert to match a work of art with someone’s taste while overcoming the main human biases is the aim of our art recommender system.
Though human preferences are highly subjective and contemporary art particularly complex to grasp, we target a high level of accuracy in our recommendations using multiple information, both textual and visual.

To exploit the visual information of our datasets, we rely on Convolutional Neural Networks (CNNs) which have demonstrated their power to analyze images, from classification tasks to style transfer. Applying them to paintings was a logical next step. First, classifying paintings' style shows surprisingly good results. The paper \cite{elgammal2018shape} explains that fine-tuned neural networks discriminate styles relatively well and display interpretations of paintings' representations. Prior to this study, such aspects have been used in \cite{elgammal2015quantifying} to investigate artistic creativity. Their approach combines proximities from visual criteria and network analysis. We note that it provides not only promising results but also a powerful tool that can be applied to much more general artistic proximities. 

Another demonstration of the power of CNNs relies on the artistic style transfer. It consists in taking two images, one as a content input and the other as a style input. The idea is to transfer the learned style into the content input image \cite{gatys2016image}. The results are very impressive and emphasize the ability of neural networks to encode the artistic style.

An important drawback of these works is the total absence of sculptures, installations, and other artistic mediums. As we will see in the article, these tools can be extended beyond paintings and are able to provide a strong analysis.

Natural Language Processing (NLP) is the field of study concerned with how to process and analyze large amounts of natural language data. Such models have proved their efficiency to analyze textual information and semantics, most notably by using neural networks. They can represent words through a low dimension vector, enabling similarity calculations.
 
These models range from a simple model like word2vec\cite{mikolov2013efficient}, that does not take into account the surrounding context, to more advanced NLP models like BERT (Bidirectional Encoder Representations from Transformers \cite{devlin2018bert}). The latter model's architecture is based on a multi-layer bidirectional Transformer encoder. One of the advantages of such model is that it is aware of the context of the words, \textit{i.e} the same word can have different representations depending on the context sentence in which it appears.

Some of the NLP applications in creative industries include Music Knowledge Discovery. In \cite{sergio2018nlp}, the authors leverage a diverse number of large corpora about a genre of music, artists biographies, reviews, etc., to derive semantic graphs between artists and entities (themes) linked through their biographies. Other applications can range from information extraction to sentiment analysis. 

In fine art, \cite{Kim2022} combined CNNs and language models to learn visual concepts from painting style, by training a BERT language model specifically for art and defining a dictionary of visual concepts with art historians.

Contemporary art content offers similar opportunities with an abundance of textual data, notably biographies, press publications and art fairs commentary. We hypothesize that combining both CNNs and NLP models enables us to attain the highest level of accuracy and robustness in our recommendation.

The most famous recommendation system is the page rank algorithm \cite{page1999pagerank}. Initially applied to rank webpages, it sets a general framework to rank elements of a graph. Recommendation systems use different technologies and can be classified into two broad groups \cite{Takaes2009}:
\begin{itemize}
    \item \textit{Content-based} approaches rely on items properties and attributes to define a similarity matrix between them. This method is well suited for use cases where user data is not yet available and can be a cold start for a more hybrid system
    \item \textit{Collaborative filtering} approaches are based on users' interactions with items. More precisely, users are recommended items that are preferred by similar users.
\end{itemize}

An example of a recommendation system applied to the art domain is CuratorNet by \cite{pablo2020curnet}. Their proposed model uses image embeddings extracted using pre-trained CNNs and trains the network for ranking with triplets $(P_u,i_+,j_-)$, where $P_u$ is the history of images preferred by a user u, whereas $i_+$ and $j_-$ are a couple of images with higher and lower preference respectively. Their model outperforms state-of-the-art VBPR \cite{he2015vbpr} as well as other simple but strong baselines such as VisRank \cite{visrank2017}. We believe that such approach will not capture the unique tastes of individual users especially in the contemporary art world where user-based recommendations tend to split users into exclusive clusters and go against the diversity we seek through our method.

In this newly digitized art ecosystem, a good amount of data about artists and artworks is publicly available while data on art collectors' behaviors are very sparse. This has driven us to focus on content-based recommendation. Furthermore, we truly believe that it provides a better interpretation of the results and is closer to what an art advisor is actually doing. Using a content-based recommendation as opposed to a collaborative filtering system brings further transparency to our users and partners \cite{Boratto2021}. With such system, we are able to tackle and monitor any bias that could emerge.

The only rare attempts of building recommendation systems for art were based solely on visual attributes of artworks. Contrary to classical art, contemporary art brings a new paradigm in our understanding of what art is since the Modern era. Such revolution resides in the fact that contemporary art goes beyond the aesthetics, the surrounding narration becomes essential where conceptualism is at the center. The taste in art no longer relies solely on its object but also on the discourse around it: the artist practice, the message conveyed in the work, the influences used by the artist and many more contextual attributes.

Our main results and contributions are:
\begin{itemize}
    \item  The creation of an exhaustive, specialized and manually annotated contemporary art dataset containing images and contextual information about artworks and artists. 
    \item Selection and training of visual and language models based on this dataset.
    \item The creation of the first contemporary art recommendation system combining visual and contextual information.
    \item Professional and exacting assessment of the performance of our recommendation system by a diverse team of art experts showed an accuracy of recommendation beyond $75\%$, compared to $11\%$ for random recommendations.
\end{itemize}

Our paper is organized as follows. First, we detail the dataset that we created and annotated in order to train and test the model. We then explain the recommendation model and its components. In the last section, we present and discuss the recommendation results we obtain.

\section{A Unique Dataset}

To conduct our study, we have created our own dataset, comprising images of artworks and contextual data about them as well as about the artists that created them. With the help of an internal team of art experts with background in contemporary art and its history, we have carefully identified the key features that encode artistic tastes from both aesthetic and contextual point of views. This encoding will be our starting point to build proximities between artists and artworks. Our approach is thus strongly supervised and requires a lot of time to manually analyze every type of information we have gathered, as we will discuss below. 

In order to have a good coverage of the contemporary artistic scene, our art specialists selected  artists and artworks that are a representative sample of what constitutes the diversity in contemporary art; including all commonly used art mediums such as paintings, installations, sculptures, photography, etc, with all relevant visual and contextual information.

\subsection{Artists}

Contextual information is a variety of elements that contribute to the understanding of an artist practice: inspirations and art movements that have influenced the artist, techniques and materials used, the themes and messages the artist is conveying through their art, etc. Building a database consists in gathering publicly available data about the artists. Our sample counts more than $1,000$ artists.
In order to increase our knowledge on artists' background, this data includes all available biographies, press articles, as well as information such as their date of birth, their working cities and of course their curriculum vitae which contains important details of past exhibitions, awards, biennials etc.

Then we define the tags used to characterize the above information through an art dictionary comprising dedicated sets of keywords. For example, the set covering the themes addressed in contemporary art, built thanks to our art experts, consists of about $350$ keywords, such as "feminism", "global warming", "spirituality", "afrofuturism" etc. Other sets describe the practices, styles, techniques, artistic mediums and materials used, backgrounds and  inspirations of the artists.
Our art experts have carefully assessed the validity of the attribution of keywords of the art dictionary and then the associations of artists with at least one keyword in common.

\begin{figure}[hbt] 
    \centering
    \subfloat[]{%
        \includegraphics[width=0.2\textwidth]{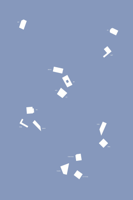}%
        \label{fig:a}%
        }%
    \hfill%
    \subfloat[]{%
        \includegraphics[width=0.2\textwidth]{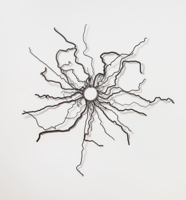}%
        \label{fig:b}%
        }%
    \hfill%
    \subfloat[]{%
        \includegraphics[width=0.2\textwidth]{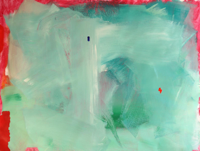}%
        \label{fig:c}%
        }%
    \hfill%
    \subfloat[]{%
        \includegraphics[width=0.2\textwidth]{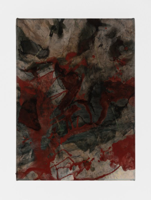}%
        \label{fig:d}%
        }%
    \caption{Abstract artworks\label{Abstract}}
    \subfloat[]{%
        \includegraphics[width=0.2\textwidth]{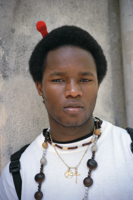}%
        \label{fig:aa}%
        }%
    \hfill%
    \subfloat[]{%
        \includegraphics[width=0.2\textwidth]{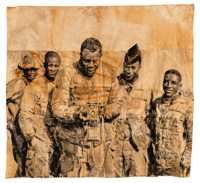}%
        \label{fig:ba}%
        }%
    \hfill%
    \subfloat[]{%
        \includegraphics[width=0.2\textwidth]{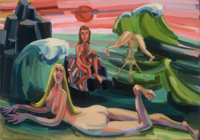}%
        \label{fig:ca}%
        }%
    \hfill%
    \subfloat[]{%
        \includegraphics[width=0.2\textwidth]{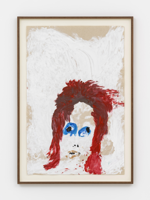}%
        \label{fig:da}%
        }%
    \caption{Figurative artworks\label{Figurative}}
    \subfloat[]{%
        \includegraphics[width=0.2\textwidth]{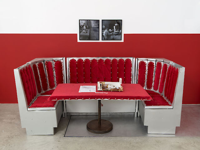}%
        \label{fig:ab}%
        }%
    \hfill%
    \subfloat[]{%
        \includegraphics[width=0.2\textwidth]{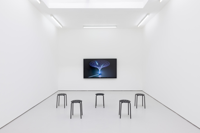}%
        \label{fig:bb}%
        }%
    \hfill%
    \subfloat[]{%
        \includegraphics[width=0.2\textwidth]{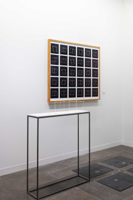}%
        \label{fig:cb}%
        }%
    \hfill%
    \subfloat[]{%
        \includegraphics[width=0.2\textwidth]{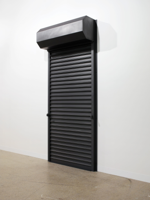}%
        \label{fig:db}%
        }%
    \caption{Conceptual artworks\label{Conceptual}}
\end{figure}

\subsection{Artworks}\label{dataset-artworks}

A possible method to represent the artworks we gathered would be to use image embeddings from a pre-trained model. However, according to our art specialists, the proximities generated using such method are very often dominated by basic descriptors such as colors, shapes and texture and ignore more subtle attributes such as artistic style, subject, etc. These descriptors are all categorical variables with multiple modalities.

Our art team gathered public images of about $5,000$ artworks spanning contemporary art in terms of time, style and practice as well as mediums (paintings, sculptures, installations, photographs, etc.) and manually annotated and classified them. The choice of the classification criteria is crucial to accurately describe an artwork from an art research perspective. We selected about $40$ categorical variables, some of them intuitive for art newcomers such as colors, artistic style and others more advanced like mark-making and themes, creating this unique dataset.

The 
figures \ref{Abstract},\ref{Figurative} and \ref{Conceptual} present a few examples of the annotation of artworks into 3 general styles: abstract, figurative and conceptual.

\section{Methodology}\label{methods}

We carefully select a training sample ($80\%$) and a test sample ($20\%$) : the first one is dedicated to train the model to annotate images of artworks with a curated set of visual attributes. We use the second sample to evaluate the annotation performance.

\subsection{Visual Neural network architecture}

Our aim is to classify the artworks using the previously defined criteria, to find deeper artistic proximities.
To do so, we consider models like AlexNet \cite{krizhevsky2012imagenet}, VGGNet \cite{simonyan2014very} and ResNet \cite{he2016deep}, which are three networks historically used in computer vision. They share common architectures ideas: smart combinations of 2D convolutional layers, pooling layers and final fully connected layers. This extracts multi-scale embeddings of the artwork images.
We use the power of transfer learning to design the classifiers we use to characterize artworks. We proceed as follows for each classification task:
\begin{itemize}
    \item We take a pretrained CNN such as AlexNet, VGGNet or ResNet.
    \item We remove the last layers.
    \item We extract the output of key intermediary layers as a feature map.
    \item We apply a technique to reduce dimension, and thus obtain a new representation of the images.
    \item We train a classification tool on top of this representation, using our dataset. We experimented with multiple methods, e.g.~various classification techniques, freezing or fine-tuning the representation, etc.
    
\end{itemize}

\begin{figure}[hbt]
    \centering
    \includegraphics[width=0.9\textwidth]{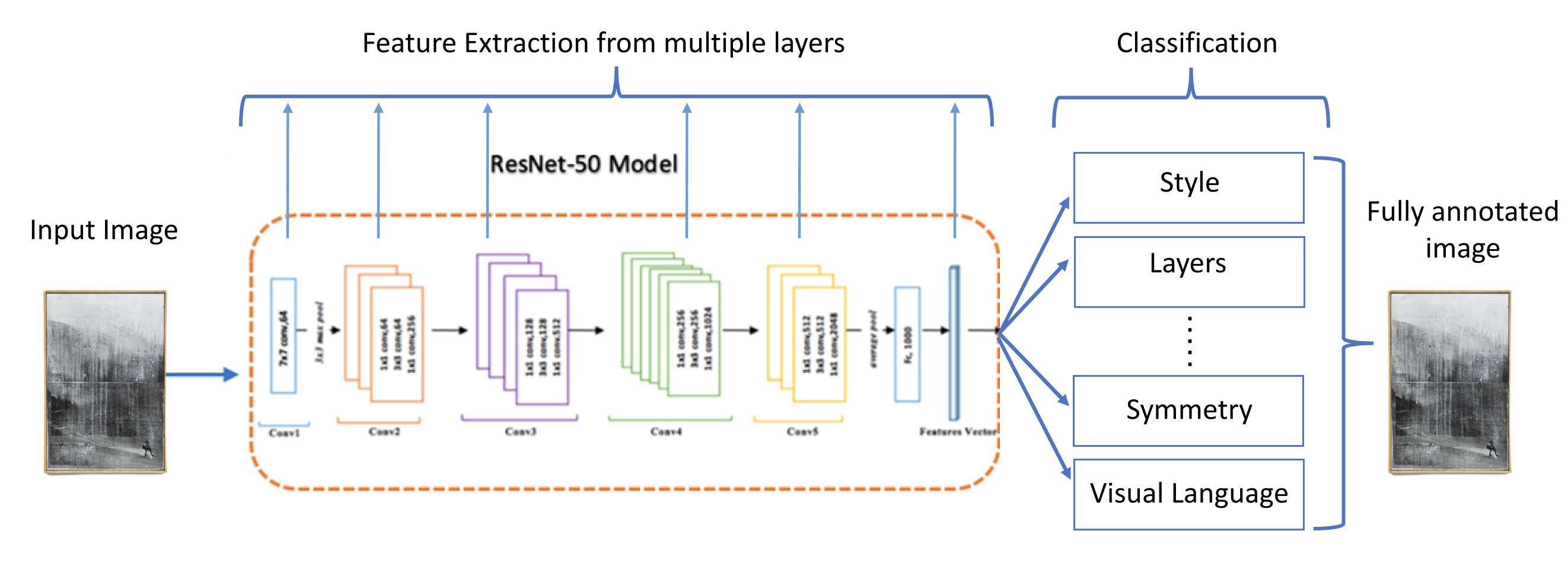}
    \caption{Classifier architecture}
    \label{architecture}
\end{figure}

We obtain a representation common to all visual criteria and a specific last layer for each classification task. This architecture is efficient to work with the relatively small sample we have. While fine-tuning the representation yielded slightly better predictions, unfreezing the pretrained model weights was also prone to over-fitting and more tricky optimization. Consequently we kept the representation frozen, and trained carefully the last layer to avoid over-fitting. 
In order to find the best classifier, we have explored common tools such as random forests, one-layer classifier (\textit{i.e.} a perceptron), linear SVM or sparse random forests and combined it with dimension reduction techniques (such as PCA). This architecture can be seen instantiated with a Resnet-50 neural network in Figure \ref{architecture}.

\begin{figure}[hbt]
    \centering
    \large
    \includegraphics[width=0.7\textwidth]{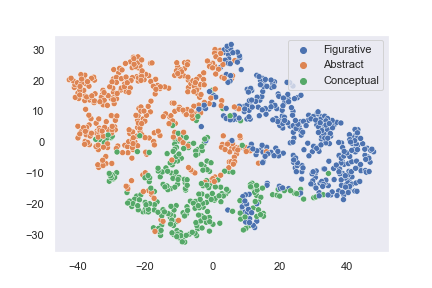}
    \caption{2D visualization of the dataset using TSNE}
    \label{clusters}
\end{figure}

In the case of a general style annotation, we can see that the neural network embeddings are relatively well-separated in Figure \ref{clusters}

\subsection{Supervised artists embeddings}

We seek to add contextual information about the artists to complement the visual information extracted from the artworks in our recommendation system. 
Our goal is to associate artistic tags from our art dictionary to artists using the analysis of textual information. We use embeddings such as word2vec and BERT to convert texts and words to vectors. As our sample is smaller than the required size to train one of these models, we start with a pre-trained network. We build an internal similarity metric between a word and a text from the cosine similarity. We use this metric to predict whether a text is associated to artistic tags. Our latest recommendation system will use an art specialised BERT model built using transfer learning and the unique artistic text data we gathered.

Then we define the full embeddings of the artists as a concatenation of the artistic tags with the information from their curriculum vitae, including the date of birth and their working cities.

We also assign different weights to different categories as not all artist attributes contribute equally to the characterisation of the artist. Some artists define themselves by their unique techniques, others by the subjects and themes of their works. 
\subsection{From the embeddings to the recommendation engine}

We define the embedding of an artwork as the concatenation of the outputs of the textual analysis of the artist and the visual analysis of the artwork. We convert this vector of categorical variables, \textit{i.e.}~classes, into a one-hot representation in order to perform efficient vector operations. We weight each variable according to the accuracy of its associated classifier. Then we compute the $L_1$ norm between artworks to build a proximity graph between artworks. 

Let us consider a graph $(m_{ij})$ where $m_{ij}$ represents the weight between the nodes $i$ and $j$. A typical example of graph-based recommendation is the page rank algorithm, which consists in letting a user jump from nodes to nodes with a probability proportional to the value $m_{ij}$ of the link. To avoid being trapped in a subsample of the graph, a small probability to visit any node is added. This mathematical trick enables this dynamic to converge toward equilibrium and then having a stationary distribution. The rank of the nodes are obtained through ranking these probabilities.

Our recommendation system functions as follows. Given an input, we rank the artworks using the graph structure of the above similarities and we return the five closest artworks under the constraint of having five different artists. More precisely, our input is an artwork made by a certain artist. Thus, we leverage two proximity graphs, one between the artists based on contextual information and the other between artworks.

\section{Results}

As a baseline, we first tested a completely random recommendation model. When the relevance of the recommended artworks was assessed by our team of art experts, the performance was $11 \%$ on average.

\subsection{Artwork annotation }

We automatically assess this procedure on the test dataset previously mentioned. The accuracy depends on the number of classes, but we compare it to the predictor that randomly selects a class with its probability of occurrences. We make sure that our estimator performs better than the random one.

\begin{figure}[h!]
    \large
    \centering
    \includegraphics[width=\textwidth]{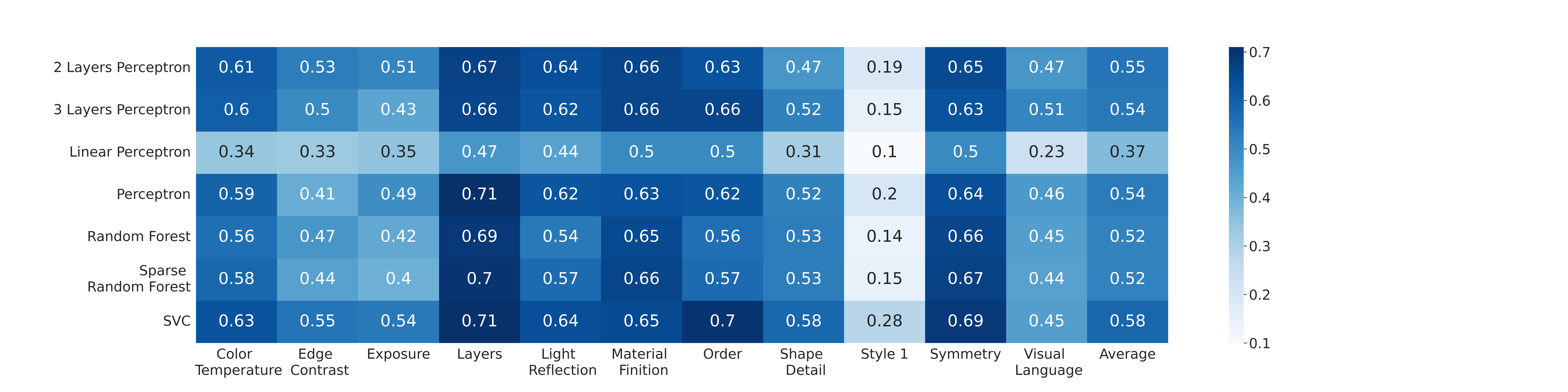}
    \caption{Classification performance benchmark on test data on a subset of artistic criteria}
    \label{performance}
\end{figure}

In Figure \ref{performance}, we can see the difference in performance between the different classifiers we tested. We chose the linear SVM classifier (SVC) for our recommendation system. 

We notice that for the style classification, the performance is very poor since we have 17 different classes in this category. To finetune the results, we merged the classes that created confusion during the classification when it also makes sense from an art expert point of view. This resulted in a real improvement showcased in Figure \ref{style}:

\begin{figure}[h!]
    \centering
    \includegraphics[width=0.3\textwidth]{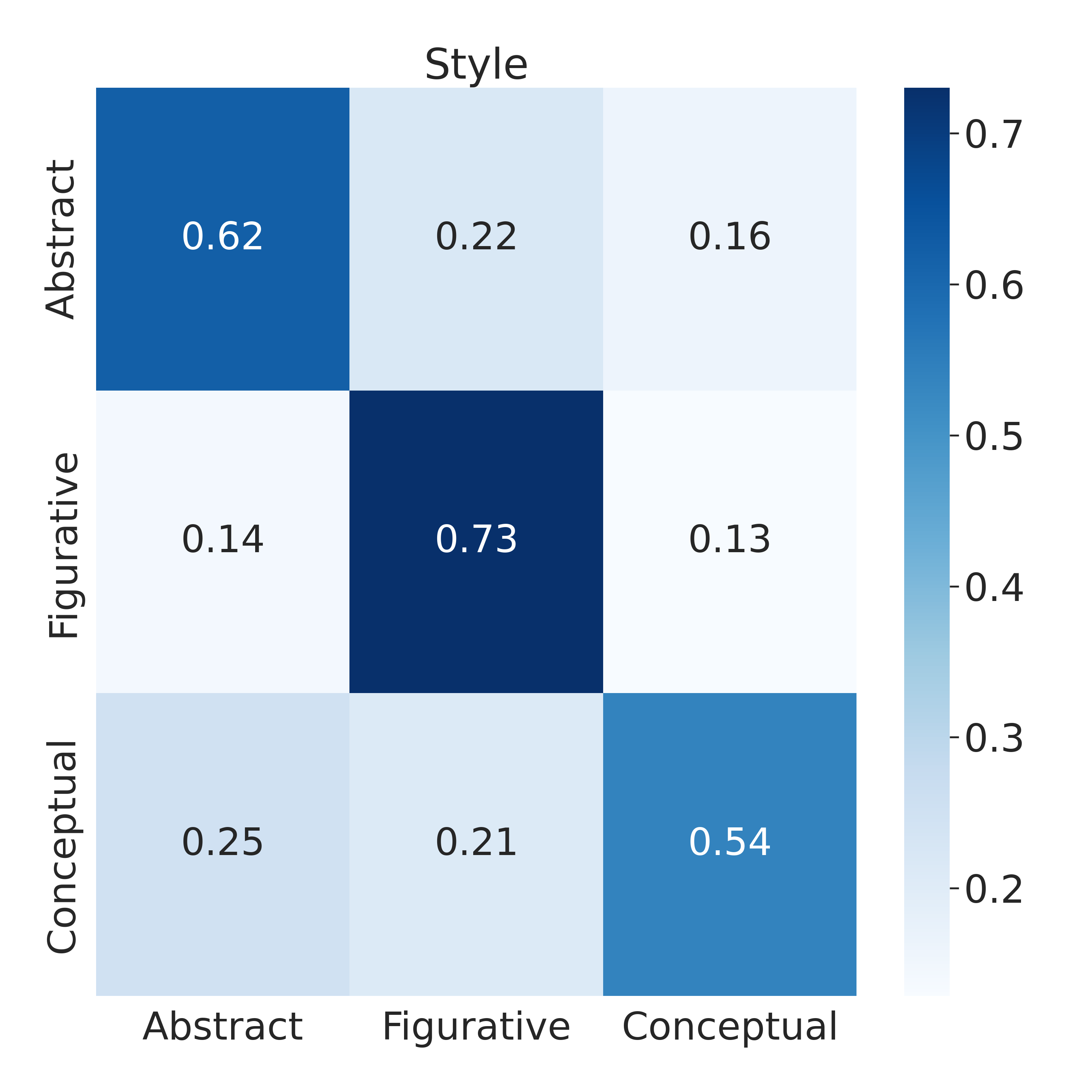}
    \caption{Confusion Matrix for the Grouped Styles with the chosen classifier}
    \label{style}
\end{figure}

\subsection{Artist tagging }

\begin{figure}[h!]
    \centering
    \includegraphics[width=0.9\textwidth]{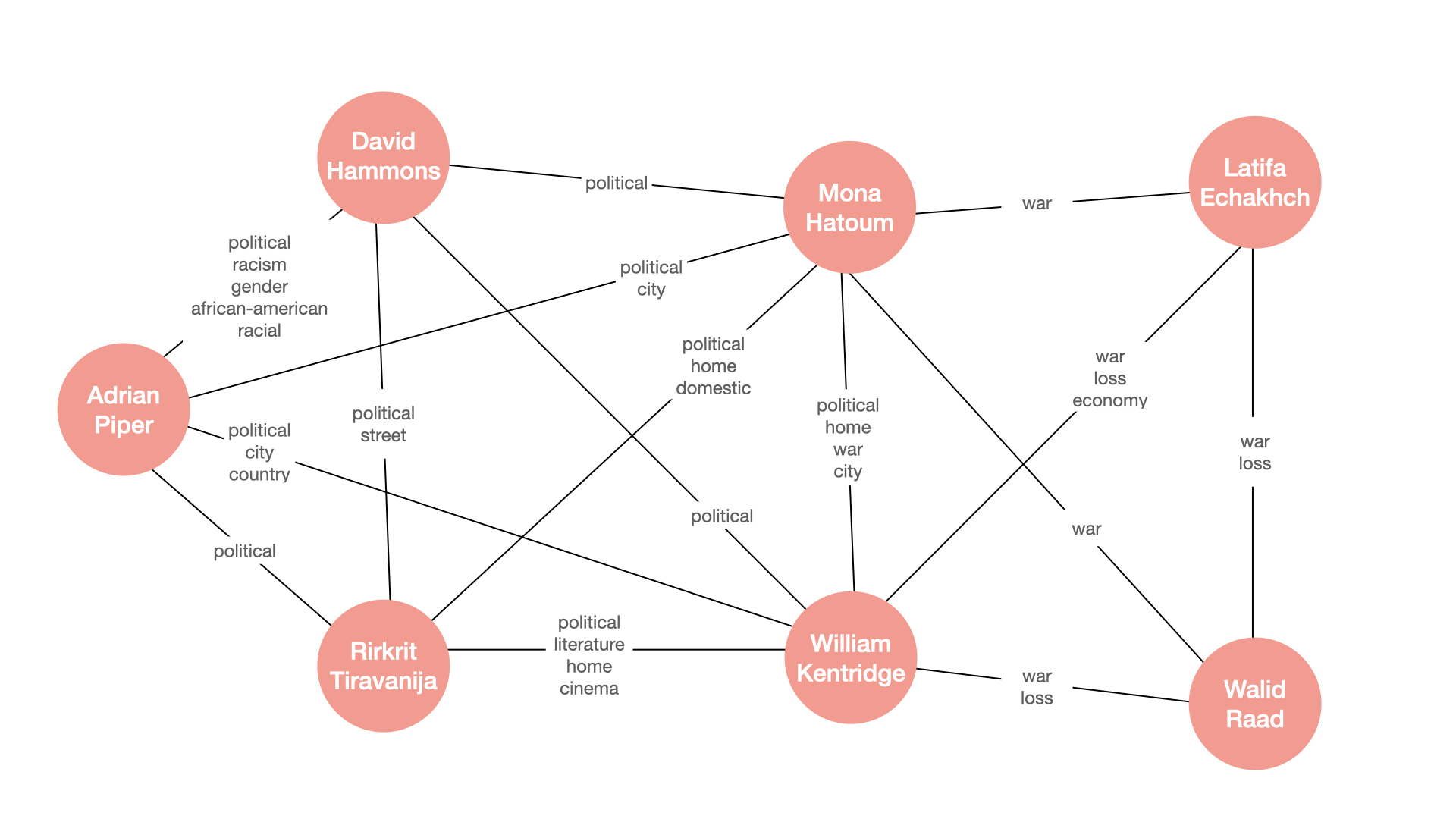}
    \caption{Subset of the artist proximity graph based on common theme tags}
    \label{graph}
\end{figure}

Our art team has carefully analysed the art dictionary tagging procedure to check its accuracy. After withdrawing misleading tags from the process, we obtain a tagging accuracy of $75\%$. Then we have also assessed if two artists with one tag in common were really linked. This leads to a score of $60\%$ when compared to our art experts' opinions and creates a proximity graph based on commonalities between artists (Figure \ref{graph})

\subsection{Recommendation}

To achieve optimal recommendation, we must define how to generate recommended artworks based on learned embeddings. Observing users' likes and dislikes of suggested artworks, some patterns emerge. For example, some users can either appreciate the visual links between artworks while others prefer the contextual links more. These trends translate into finding an optimal weight to combine the two types of embeddings, to maximize the precision score, which is the proportion of good recommendations.

\begin{figure}[tbh] 
    \centering
    \captionsetup[subfloat]{labelformat=empty}
    \subfloat[]{%
        \includegraphics[width=0.15\textwidth]{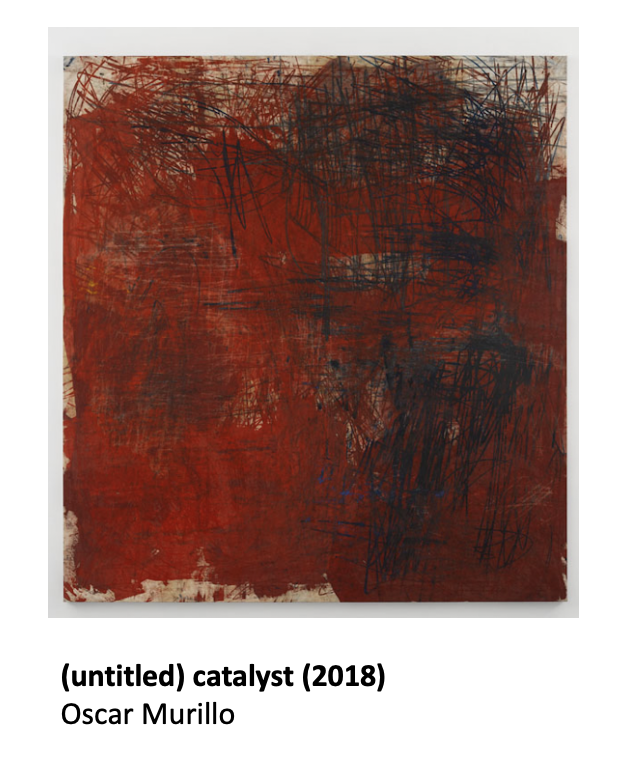}%
        }%
    \hfill%
    \subfloat[]{\raisebox{10mm}{%
        \includegraphics[width=0.05\textwidth]{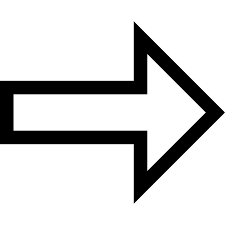}%
        
        }}%
    \hfill%
    \subfloat[]{%
        \includegraphics[width=0.15\textwidth]{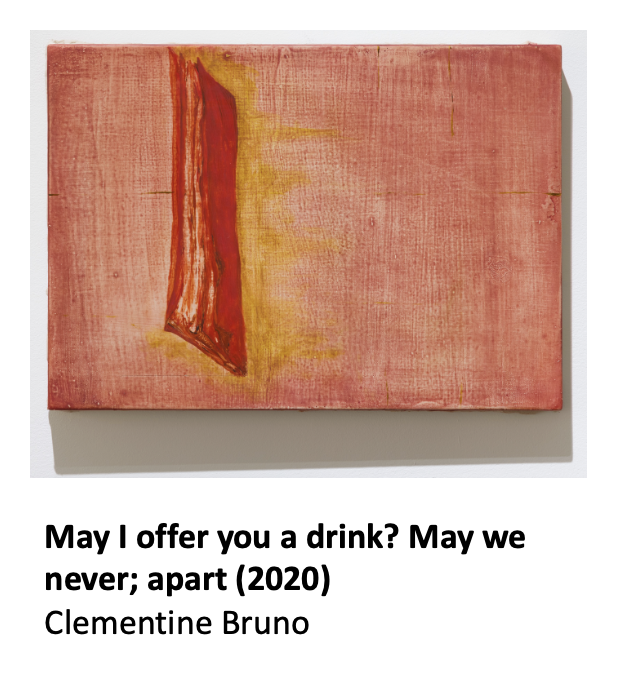}%

        }%
    \hfill%
    \subfloat[]{%
        \includegraphics[width=0.15\textwidth]{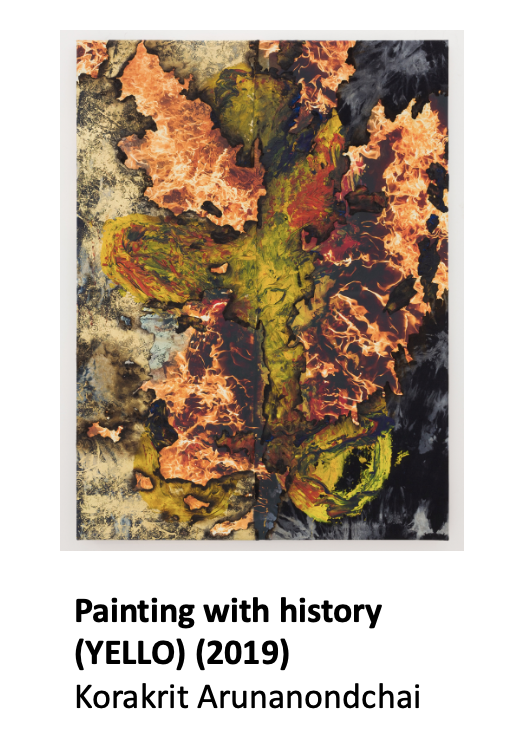}%

        }%
    \hfill%
    \subfloat[]{%
        \includegraphics[width=0.15\textwidth]{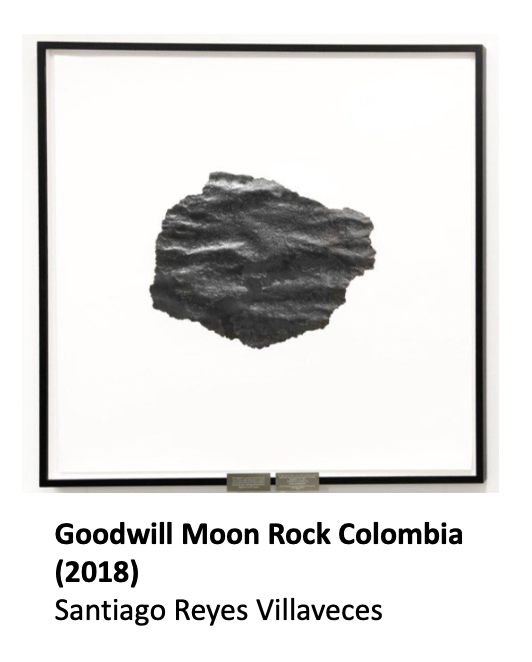}%

        }%
    \hfill%
    \subfloat[]{%
        \includegraphics[width=0.15\textwidth]{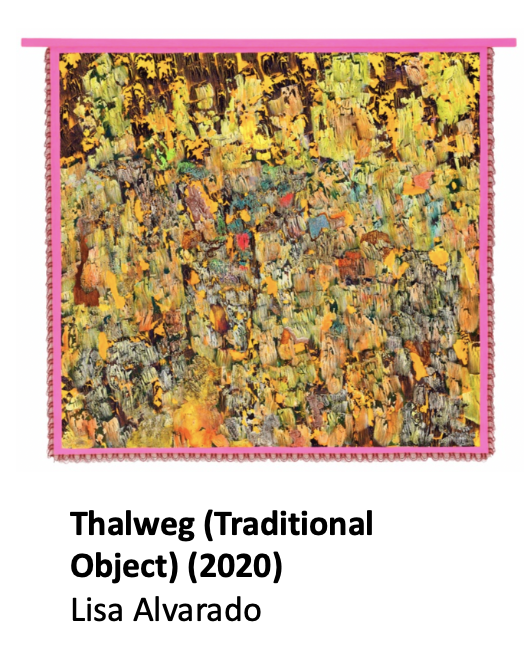}%

        }%
    \hfill%
    \subfloat[]{%
        \includegraphics[width=0.15\textwidth]{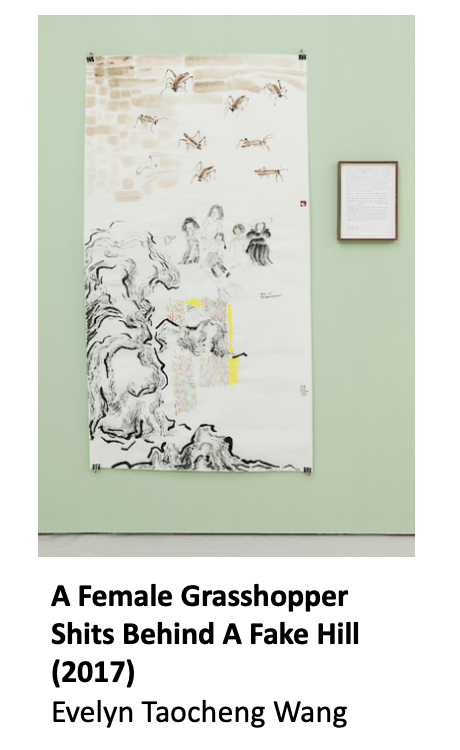}%

        }%
    \caption{Visual Recommendation\label{Visual}}
    \captionsetup[subfloat]{labelformat=empty}
    \subfloat[]{%
        \includegraphics[width=0.15\textwidth]{figures/recommendations_bis/murillo_.png}%
        }%
    \hfill%
    \subfloat[]{\raisebox{10mm}{%
        \includegraphics[width=0.05\textwidth]{figures/arrow.png}%
        
        }}%
    \hfill%
    \subfloat[]{%
        \includegraphics[width=0.15\textwidth]{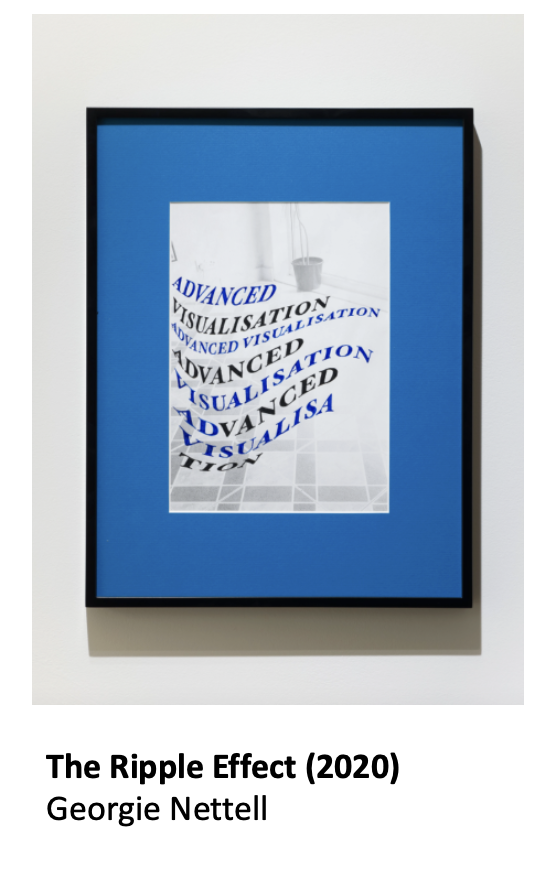}%
        }%
    \hfill%
    \subfloat[]{%
        \includegraphics[width=0.15\textwidth]{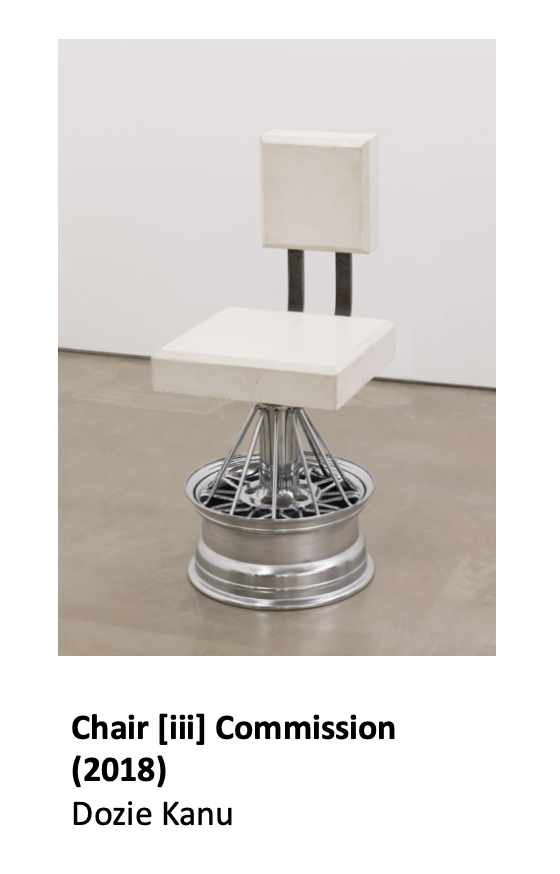}%

        }%
    \hfill%
    \subfloat[]{%
        \includegraphics[width=0.15\textwidth]{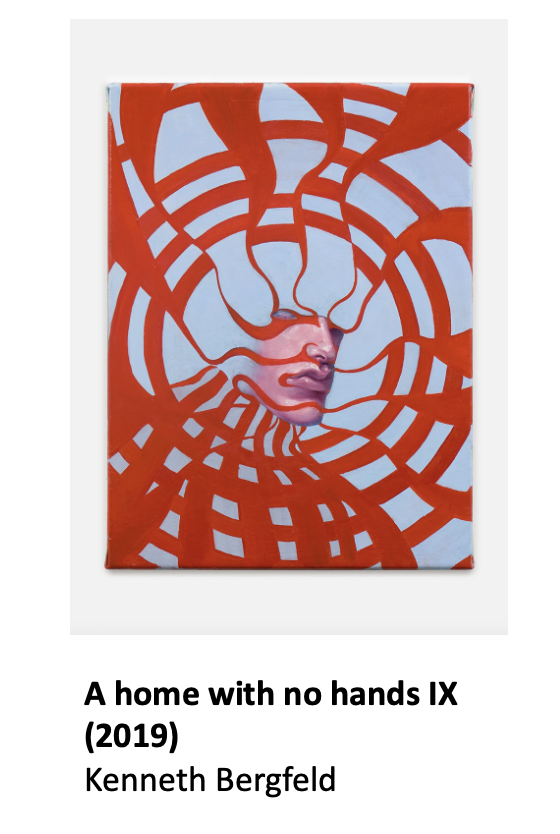}%

        }%
    \hfill%
    \subfloat[]{%
        \includegraphics[width=0.15\textwidth]{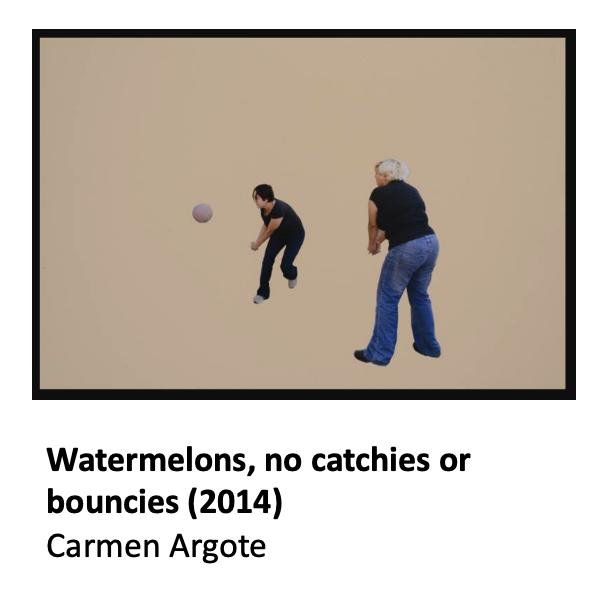}%

        }%
    \hfill%
    \subfloat[]{%
        \includegraphics[width=0.15\textwidth]{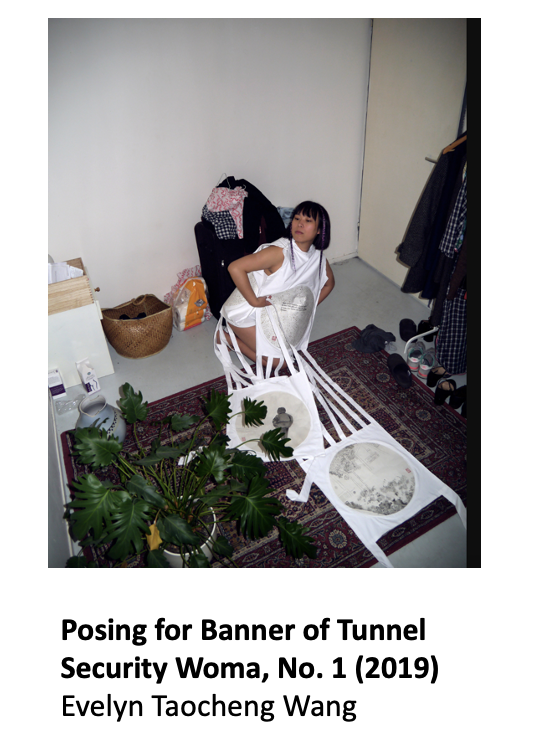}%

        }%
    \caption{Contextual Recommendation\label{Contextual}}
    \captionsetup[subfloat]{labelformat=empty}
    \subfloat[]{%
        \includegraphics[width=0.15\textwidth]{figures/recommendations_bis/murillo_.png}%
        }%
    \hfill%
    \subfloat[]{\raisebox{10mm}{%
        \includegraphics[width=0.05\textwidth]{figures/arrow.png}%
        
        }}%
    \hfill%
    \subfloat[]{%
        \includegraphics[width=0.15\textwidth]{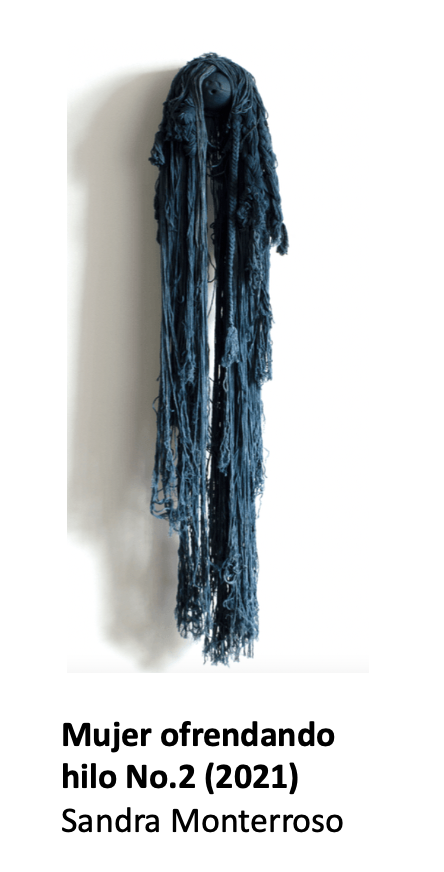}%

        }%
    \hfill%
    \subfloat[]{%
        \includegraphics[width=0.15\textwidth]{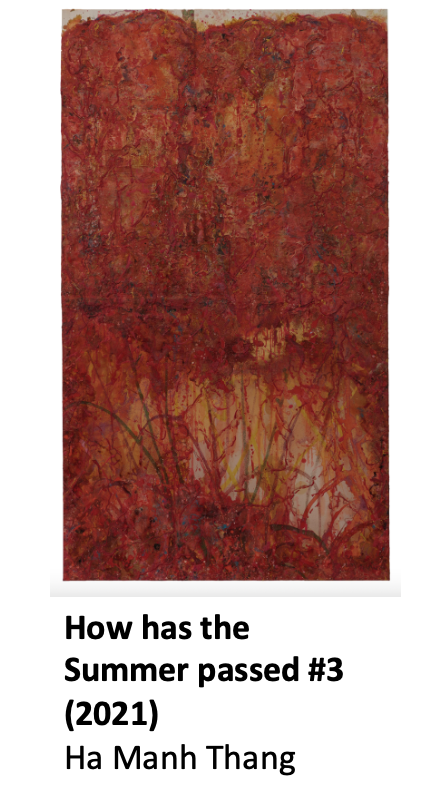}%

        }%
    \hfill%
    \subfloat[]{\raisebox{5mm}{%
        \includegraphics[width=0.2\textwidth]{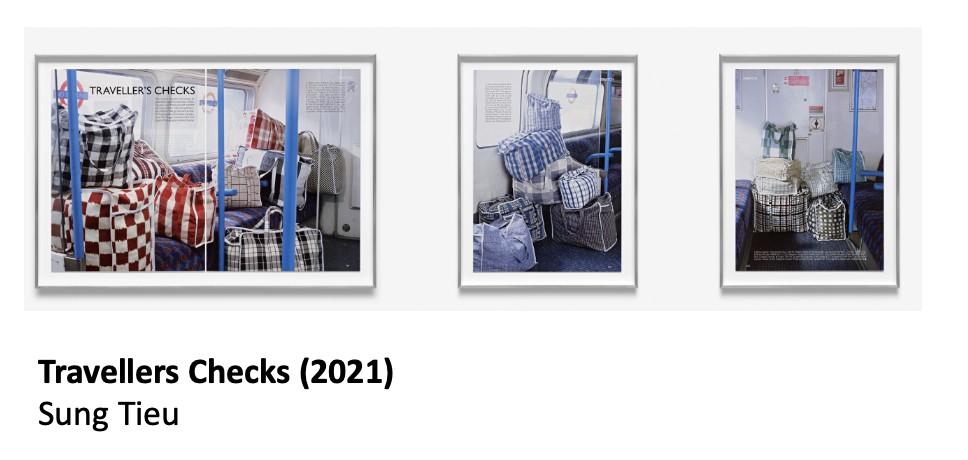}%

        }}%
    \hfill%
    \subfloat[]{\raisebox{5mm}{%
        \includegraphics[width=0.15\textwidth]{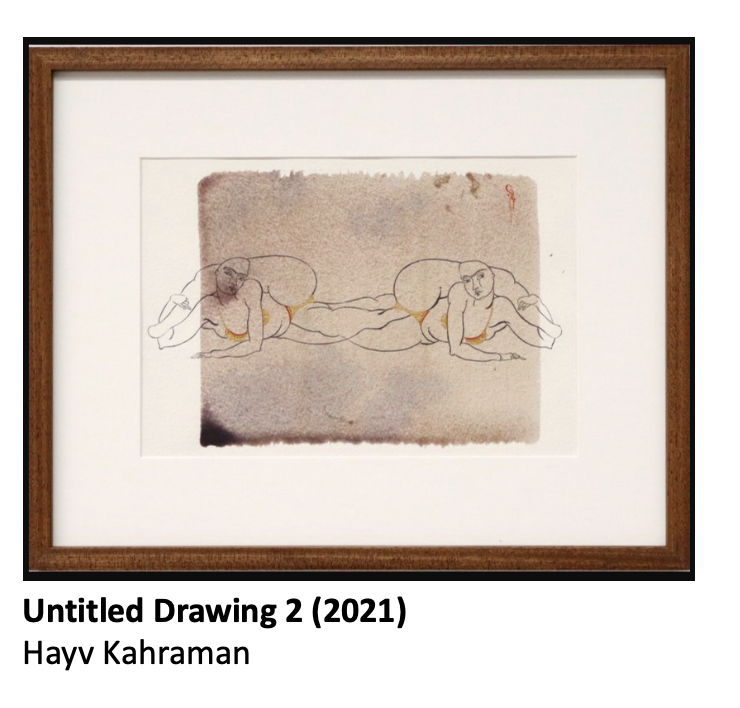}%

        }}%
    \hfill%
    \subfloat[]{\raisebox{5mm}{%
        \includegraphics[width=0.15\textwidth]{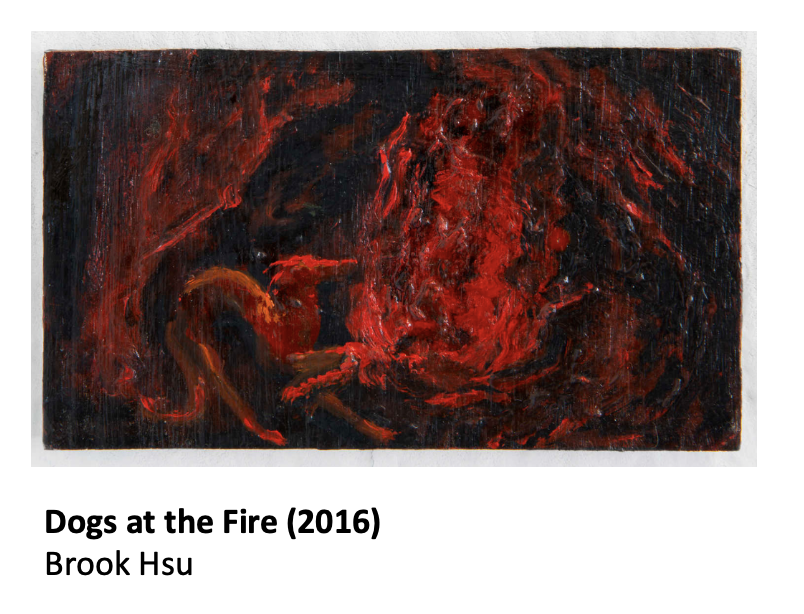}%

        }}%
    \caption{Final Recommendation\label{reco}}
\end{figure}

To ascertain the added value of our model, we compare the recommendations obtained using only the visual model ($65 \%$), the purely contextual recommendation ($60 \%$)  and the final weighted recommendation ($75 \%$). In Figures \ref{Visual},\ref{Contextual} and \ref{reco} we give an example of each recommendation from the same input (Oscar Murillo, (untitled) catalyst, 2018).

The challenge was to generate relevant art recommendations not only for art aficionados but also to the eyes of recognized experts. Therefore, we submitted the results to seven art experts with a background in art history and acute knowledge of the contemporary art scene, giving us a lower bound of the precision. Our recommendation engine demonstrates good performances to match artworks from a selection with an average performance of $75 \%$, after a fine-tuning of our custom weights of variables. It appears that giving greater weights on contextual than visual variables is more appreciated by our experts. Indeed, it is able to capture much more subtle links between artworks and artists that cannot currently be caught by the eyes of our neural networks. Furthermore, having an accurate numerical value of the distance between artists and artworks allows us to recommend art beyond the echo chamber of the users' tastes and introduce them to new artists and styles, by suggesting some farther artworks. A possible improvement could be enriching the set of visual tags with ones more related to art history or context to bridge the gap between visual aspects and contextual data.

\section{Conclusion}

In this work we showed the potential for a recommendation system to use the available information and present outputs in line with the opinions of art experts. To achieve this, we created a unique and proprietary dataset of artworks and artists tagged with advanced artistic information and used it to train multiple models that we combined to create proximity graphs. This enabled us to suggest relevant recommendations from one artwork as input and achieve great accuracy without the need for user data. 

Moreover, as the user's journey evolves and more interactions with the recommendation accumulate each day, we are able to define a set of attributes to characterize the taste profile of a given user. This allows us  to refine even more the recommendation to the unique taste of each user by changing the weights of the different components of the system. In the future, we would like to build recommendations not only from a single artwork but also from an evolving set of artworks. The musical analogy would be to build a recommendation from a playlist. Our first results set the basics to create such tools. Furthermore, as our dataset gets bigger and better, we will train CNN and NLP models specific to contemporary art for better performance and more granular understanding of artistic links and concepts. 

More broadly, we think such a content-based model that combines multiple types of data can be applied in many ways to extract nuanced information, whether to recommend, classify or automatically annotate large amounts of data. Studying how such a recommendation model can avoid the pitfalls of collaborative filtering models, such as popularity bias and lack of diversity that can occur with users trapped in echo chambers of homogeneous taste\cite{Werner2020}, presents great opportunities for future research.

\section{Acknowledgments}

We truly thank the whole Docent team for the useful discussions and commentary, as well as the galleries for gracefully agreeing to provide us with the artworks displayed in this paper.


\bibliographystyle{alpha}
\bibliography{bibliography}

\end{document}